\documentclass[11pt]{scrartcl}
\usepackage{subcaption}
\usepackage{amsmath}
\usepackage{natbib}
\usepackage{xcolor}
\usepackage{graphicx}
\usepackage{hyperref}
\usepackage{verbatim}
\usepackage[inline]{enumitem}
\usepackage{booktabs}
\usepackage{todonotes}
\usepackage[normalem]{ulem}
\usepackage[inline]{enumitem}
\presetkeys%
    {todonotes}%
    {inline}{}

\presetkeys%
    {todonotes}%
    {inline}{}
\usepackage{natbib}
\setcitestyle{authoryear}

\usepackage{geometry}
\newgeometry{vmargin={20mm}, hmargin={25mm,25mm}} 

\usepackage{fancyhdr}
\usepackage{caption}
\usepackage{xcolor}

\title{Conversational Recommendation: \\A Grand AI Challenge}
\author{
Dietmar Jannach\\
                \mbox{\small University of Klagenfurt, Austria}\\
                \mbox{\small dietmar.jannach@aau.at}\\ \and
        Li Chen\\
                \small Hong Kong Baptist University, Hong Kong\\
                \small lichen@comp.hkbu.edu.hk\\
}
\date{\today}
\date{~}

\begin{document}

\maketitle

\begin{abstract}
Animated avatars, which look and talk like humans, are iconic visions of the future of AI-powered systems. Through many sci-fi movies we are acquainted with the idea of speaking to such virtual personalities as if they were humans. Today, we talk more and more to machines like Apple's Siri, e.g., to ask them for the weather forecast. However, when asked for recommendations, e.g., for a restaurant to go to, the limitations of such devices quickly become obvious. They do not engage in a conversation to find out what we might prefer, they often do not provide explanations for what they recommend, and they may have difficulties remembering what was said one minute earlier. Conversational recommender systems promise to address these limitations. In this paper, we review existing approaches to build such systems, which developments we observe today, which challenges are still open and why the development of conversational recommenders represents one of the next grand challenges of AI.\footnote{Paper accepted for publication in AI Magazine (2022).}
\end{abstract}

\section*{Introduction}
We are increasingly used to interact with machines in natural language. We ask Siri to find us a nearby Italian restaurant, we request Alexa to play some music, and we even talk to our cars, instructing them to route us to our destination. Many of us have also interacted with chatbots, which are used by companies as a first contact point for customer service. Being able to interact with machines that are able to converse like humans has long been an iconic vision of the future in sci-fi movies. With the recent advances in natural language processing (NLP), the availability of huge pretrained language models like GPT-3, and the progress of machine learning in general, one may therefore think we are already close to achieving this vision.

Looking closer at the above-mentioned interactions, we observe that the devices we talk to are often good at reacting to individual commands. However, these systems sometimes reach their limits in situations when a multi-step conversation would be required to achieve a task. A typical example is the reservation of a flight after discussing different options with a travel agent. Another common problem setting, where typically more than a one-shot interaction is needed, is that of recommending something based on individual short-term user preferences. This process is commonly referred to as \emph{conversational recommendation}, cf. \citep{Jannach2020ASO}, and such problems are the focus of this paper.

Conversational recommendation is in several ways different from the conventional, non-interactive presentation of item suggestions as found, e.g., on the landing page of Netflix. First, while conventional recommendations rely on \emph{push} communication, conversational recommender systems (CRS) support multi-turn and mixed-initiative interaction patterns. Moreover, in particular natural language based CRS combine aspects of \emph{recommendation} and \emph{search}, i.e., they allow users to make queries. Generally, CRS in many cases target at problem settings where no long-term user profiles are available and where user requirements and the user's context are interactively acquired.

Interactive and conversational recommendation on the Web has been explored in the academic literature since the mid-1990s, see \citep{hammond1994findme} or \citep{LindenG-UM97} for early \emph{critiquing}-based approaches. In these early systems, an interaction model was implemented where the system made recommendations on which users could apply feedback in the form of critiques with respect to certain item attributes. For example, users could state that they were looking for something ``cheaper'' or for a laptop of ``higher processor speed''.
While these systems enabled interactive dialogues, the number of supported interaction types was limited and often pre-defined based on domain knowledge. Later on, the first conversational recommender systems to support a natural language interface were proposed \citep{Thompson:2004:PSC:1622467.1622479}. These early systems were however often hampered by the limited capabilities of NLP technology available at that time. Since then, NLP technology has greatly improved and represents one main pillar of academic CRS that are implemented as chatbots \citep{iovine2020conversational,qiu2017alime}. In the majority of today's real-world chatbot applications, the system's responses are however mostly based on pre-defined templates, and the main
tasks where AI technology comes into play, besides language processing, include the detection of the user's intents and the recognition of entities in the user utterances. To avoid the bottleneck that comes with the definition of the templates, recent \emph{end-to-end} learning systems commonly train complex models on large corpora of recorded recommendation dialogues between humans.
Table \ref{tab:applications} lists a number of conversational recommenders that were successfully applied in various domains over the years.

\begin{table}[t]
\scriptsize
\centering
\caption{Example applications of conversational recommender systems over time}
\label{tab:applications}
\begin{tabular}{p{2.5cm}p{2.5cm}p{9.7cm}}
\hline
\textbf{System} & \textbf{Interaction Modality} & \textbf{Description} \\
\hline
Wasabi Personal Shopper \citep{Burke:1999:WPS:315149.315486} & Forms  \& Buttons
& The Wasabi Personal Shopper was a database browsing tool based on case-based reasoning and the earlier FindMe systems. It allows users to critique a recommended product so that the system can return a new product based on their feedback.
\\  \hline
\textsc{Advisor} \textsc{Suite} / CWAdvisor  \citep{Jannach:2004:ASK:3000001.3000153} & Forms  \& Buttons
&  \textsc{Advisor} \textsc{Suite} was a commercialized software for personalized sales advisory following a knowledge-based approach. Several applications built with the CWAdvisor were deployed in the domains of electronics, investment products, wine, and fine cigars.
\\  \hline
MobyRek / STS \cite{DBLP:journals/expert/RicciN07,DBLP:conf/ecweb/BraunhoferER14} & Forms \& Buttons on Mobile & MobyRek was one early CRS designed for mobiles, and it supported critiquing for recommending travel-related items. South Tyrol Suggests (STS) is a more recent smartphone application able to leverage context information for the recommendation of Points-Of-Interest.
\\  \hline
AI bot for shopping \citep{DBLP:conf/aaai/YanDCZZL17} & Natural Language & This AI bot is a task-oriented dialogue system built as a shopping assistant for online mobile e-commerce. It is one of the first Chinese AI bots used in online shopping environments with millions of consumers. It leverages various knowledge sources and crowdsourcing to deal with large catalog size and cold-start issues.
\\  \hline
AliMe Chat \citep{qiu2017alime} & Natural Language & AliMe Chat is an open-domain industrial chatbot relying on a hybrid end-to-end deep learning approach. It was integrated and tested within a real-world intelligent assistant in the e-commerce domain at Alibaba to support customer service, shopping guidance, and life assistance such as flight booking.
\\  \hline
\end{tabular}
\end{table}

Given these success stories and the recent technical developments, one might think that building a CRS is a solved problem. Building a CRS also seems much easier than creating a \emph{general} conversational AI system because \emph{(i)} only one specific task, i.e., recommendation, has to be supported and \emph{(ii)} many field-tested machine learning models exist to determine item recommendations for a given user profile. However, an analysis of two very recent learning-based approaches indicated that today's academic systems are far from being usable in practice \citep{JannachManzoor2020}. These systems were found to be rather limited in terms of what kind of conversational acts they support or in terms of their ability to maintain the dialogue context. A simulation based on dozens of dialogues ultimately revealed that the system responses were considered to be meaningful by human evaluators in less than two third of the cases, raising questions about their applicability in practice.

Engineered CRS, which are based on pre-defined templates or explicitly coded dialogue knowledge, are often able to avoid such problems as their behavior is predictable. Ultimately, however, our goal is to avoid the knowledge engineering effort required by such systems and to build systems that are able to learn to converse from data, at least to a certain extent.
In this paper, we discuss why building such learning-based CRS, which are able to engage in a recommendation dialogue at the level that we would expect from a human, is one of today's grand challenges and testbeds for AI-powered technologies.
We first discuss the various aspects that make building a CRS difficult. We then review limitations of current approaches in particular with respect to how we evaluate systems in academia. Finally, we provide an outlook on possible next steps for building next-generation conversational recommenders.

\section*{Why Building a CRS is Difficult}
\label{sec:definition-and-tasks}
A conversational recommender system can be characterized as ``\emph{a software system that supports its users in achieving recommendation-related goals through a multi-turn dialogue''} \citep{Jannach2020ASO}. A recommendation-related goal in that context can, for example, be to help users find relevant items or, more generally, to make better decisions. However, there can also be more indirect goals like helping users to understand the space of options or explaining to them why a certain option is a good choice for them.

Following this definition, a CRS is a task-oriented system. This differentiates CRS from general conversational AI systems, including the famous ELIZA system from the 1960s. In some ways, building a CRS might therefore appear to be an easier task, because the conversation to be supported is usually bounded to a few of pre-defined tasks and dialogue situations. Moreover, the competence of a particular CRS can furthermore be limited to a certain domain, e.g., movies.

However, achieving a certain naturalness of conversational recommendation dialogues can be challenging. For example, in case of a CRS supporting natural language interactions, the virtual CRS agent should probably be able to respond to chit-chat (``phatic'') user utterances. Furthermore, a conversation between humans---which a CRS might aim to mimic---is much richer than just answering questions like ``\emph{What is a good sci-fi movie?}''. In such a conversation, the initiative might also switch between dialogue partners, thus requiring a system that supports both \emph{user-driven}, \emph{system-driven}, or \emph{mixed-initiative} dialogues. Moreover, the system must be able to respond to a variety of possible \emph{user intents}, e.g., providing or revising preference information, asking for explanations, or rejecting a recommendation. Finally, the CRS must be able to keep track of the ongoing dialogue and possibly even past interactions with the user, as done in \citep{Thompson:2004:PSC:1622467.1622479} or \citep{DBLP:journals/expert/RicciN07}. We discuss these challenges in more detail next after we review the main building blocks of a CRS.

\subsection*{Conceptual Architecture of a CRS} 
\label{subsec:conceptual-architecture}
\begin{figure}
  \centering
  \includegraphics[width=.9\textwidth]{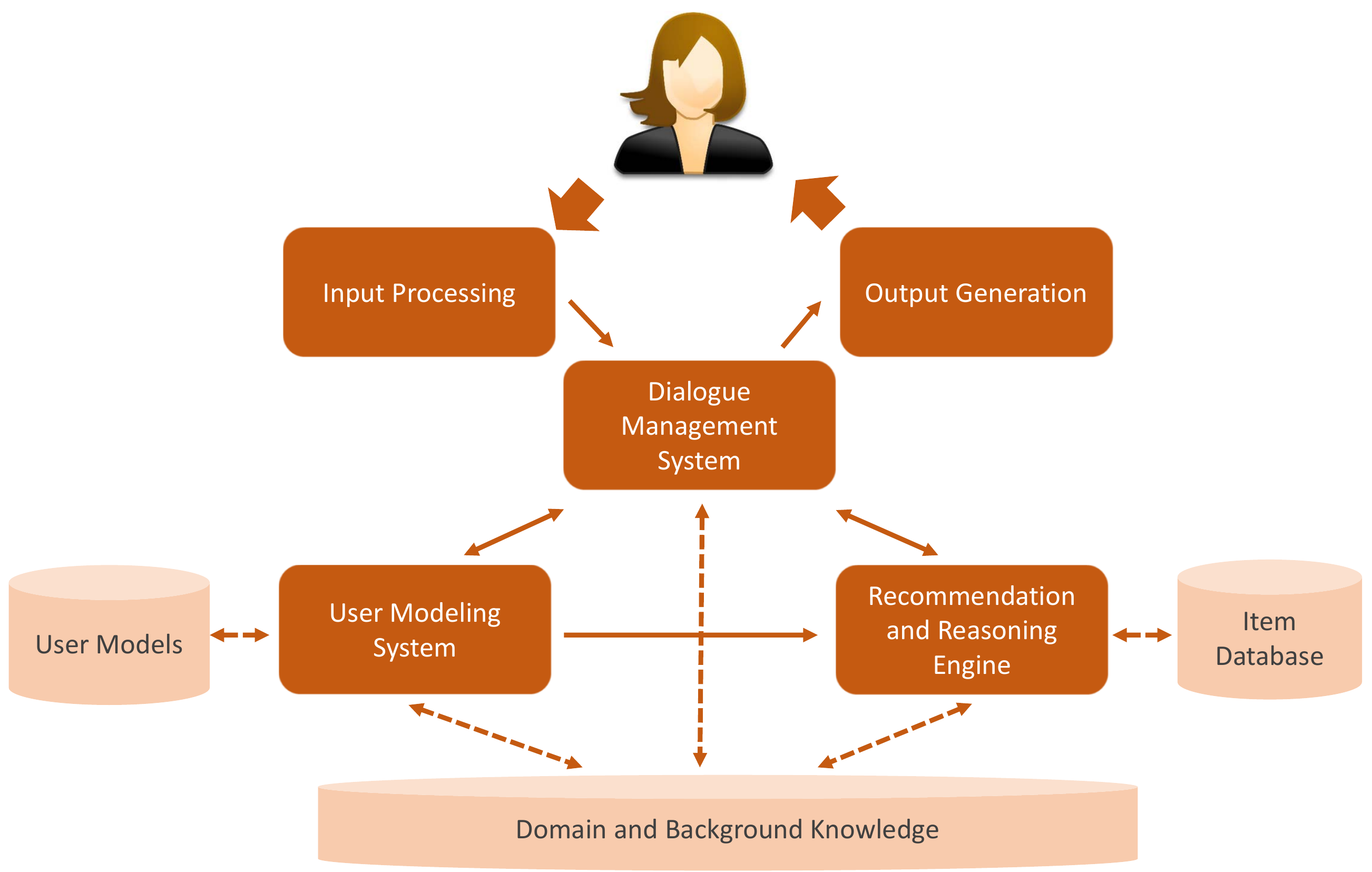}
  \caption{Conceptual architecture of a CRS (adapted from \cite{Jannach2020ASO})}\label{fig-architecture}
\end{figure}

Any CRS is an interactive software application, which repeatedly processes inputs of end users and reacts in appropriate ways, e.g., by making a recommendation, by asking questions about preferences, or by providing explanations. To support such interactions, the conceptual architecture of a CRS typically comprises the components sketched in Figure \ref{fig-architecture}.

\paragraph{Input and Output Processing} The interaction with a CRS can be designed in various ways and support different \emph{modalities}. In traditional critiquing systems, users typically interact with the system through pre-defined forms, either on the web or on the mobile. This is the most simple form of \emph{input processing} as the provided screen elements and their semantics are defined when the system is designed. More recent systems, however, also aim to support conversations in natural language, either in spoken or written form. Implementing such a functionality leads to increased complexity and requires the inclusion of additional system components, e.g., for speech-to-text conversion or for named-entity recognition (the latter requiring an underlying knowledge of known entities). In terms of the system \emph{outputs}, the traditional approach is to present the next interaction screen to the user, which may contain various interaction elements (e.g., radio buttons for preference refinement) or a table of item suggestions. In cases where natural language interaction is supported, additional components, e.g., for text-to-speech conversion, are required.

Even more complexities can arise when more than one interaction modality should be supported. On the input side, it is, for example, not uncommon in chatbot applications that users in some situations are asked to operate pre-defined interactive elements such as buttons, and at the same time are allowed to interact with the system by typing free text. One of the difficulties in such a design is to understand which is the most suitable interaction form in a given situation \citep{iovine2020conversational}. In other applications, a \emph{hybrid} design of the input and output modalities might be desired, where the input is for example speech-based, but the output is a combination of speech output, structured textual and visual information. Such a hybrid system might in particular be favorable when more than one recommendation should be presented to the user through a multi-modal user interface. A CRS on a smartphone, for example, might accept voice input in the preference elicitation phase, and then return the best recommendation via voice output. Finally, several approaches exist that rely on \emph{embodied conversational agents} to provide a more persuasive or emotional user experience and to thereby increase the effectiveness of the system, see, e.g., \citep{Foster2010umuai}.

\paragraph{User Modeling and Recommendation Reasoning Components}
On the \emph{backend} side of a CRS, multiple components are usually needed. These include, as a central element, a component that generates recommendations that match the user's preferences. These preferences are maintained with the help of a user-modeling component. This component may either consider only the user's preferences stated in the ongoing dialogue, or it may be able to also make use of long-term preference profiles \citep{DBLP:journals/expert/RicciN07,Thompson:2004:PSC:1622467.1622479}. How to combine long-term and short-term preferences---which may be changing and thus conflicting across sessions---in the best way, is so far a largely unexplored question.

From a technical perspective, various approaches are possible to generate recommendations. One can, e.g., rely on explicit recommendation rules and constraints, as done in critiquing or constraint-based systems \citep{Jannach:2004:ASK:3000001.3000153,chenpucritiquing2012,recsyshandbook2015constraints}. Other approaches train machine learning models based on collaborative information (e.g., rating datasets), which is then combined with preference information from the ongoing dialogue \citep{christakopoulou2016towards} and maybe structured external data. Besides the generation of recommendations, the backend components might furthermore support additional reasoning tasks, for example, the computation of a user-oriented explanation for the recommendation. Also, it might implement complex reasoning functionality, e.g., in constraint-based systems, to determine recommendations in case the customer's requirements cannot be fulfilled by any recommendable item.

\paragraph{Background Knowledge}
To accomplish these tasks, the backend uses different types of knowledge. Besides an optional database containing long-term preference information, a CRS has at least explicit knowledge about the items that can be recommended, sometimes including meta-data. Various other types of knowledge can be integrated as well. This may include explicit knowledge about possible dialogue states and  supported user intents as well as different forms of ``background knowledge''. The background knowledge can include both additional item-related knowledge and structured world knowledge (e.g., from DBPedia), as well as unstructured textual sources such as logs of recorded recommendation dialogues between humans.

This latter class of background information, i.e., logs of human-to-human conversations, often serves as a main basis for end-to-end learning approaches, as in \citep{chen-etal-2019-towards} and \citep{li2018towards}. In the last few years, a number of such dialogue corpora were published, such as the ReDial dataset \citep{li2018towards} that consists of over 10,000 dialogues created with the help of crowdworkers.

An alternative way of collecting such recommendation dialogue corpora with crowdworkers is the use of Wizard-of-Oz studies. In such studies, the participants interact with a human agent without knowing that it is not a chatbot. \cite{radlinski2019coached} used such a study to build a dataset focused on the preference elicitation process. Furthermore, a CRS might also rely on more \emph{general} dialogue corpora, i.e., ones that are not necessarily centered around a recommendation task. Facebook, for example, released different datasets in the context of the \emph{bAbI} project, designed to serve as a basis to train end-to-end learning systems for different tasks, e.g., for restaurant bookings. Finally, recent works explore the use of very general pretrained language models like BERT or GPT-3 in the context of conversational recommender systems. \cite{WhatDoesBertKnow2020} for example explore what such a general model like BERT already knows about movies, e.g., about their genres, and discuss where such models currently succeed and where they fail.

\paragraph{Dialogue Management}
Today's voice controlled digital assistants often fail to maintain the context of an ongoing conversation and, for example, do not remember or take into account what was said in previous interaction cycles. Keeping track of the conversation is however a central feature in a CRS, e.g., when the system tries to interactively acquire the user's needs or preferences.

\begin{figure}
  \centering
  \includegraphics[width=.4\textwidth]{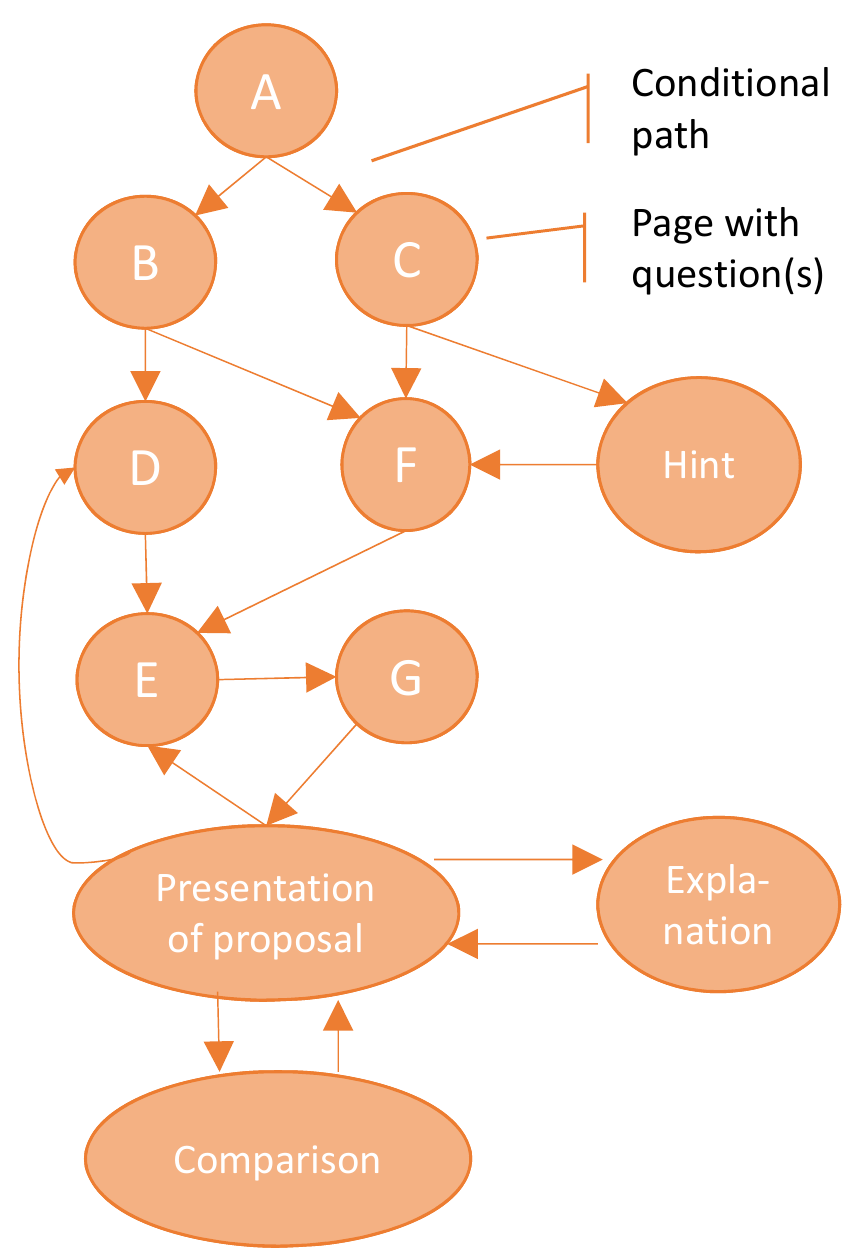}
  \caption{State transition graph (adapted from \cite{Jannach2020ASO})}\label{fig-atate-graph}
\end{figure}

To keep track of the current situation, CRS typically rely on a pre-defined set of \emph{dialogue states}. These states can either be defined explicitly, e.g., in the form of a state transition graph \citep{DBLP:journals/expert/RicciN07,Jannach:2004:ASK:3000001.3000153}, or implicitly. Figure \ref{fig-atate-graph} shows a dialogue graph that determines not only the pre-defined states, but also the possible transitions, i.e., the \emph{conversational moves}. Encoding the knowledge in this way is common in critiquing and  ``slot-filling'' preference elicitation approaches, where the system asks questions to the user about her preferences for a pre-defined set of item attributes. Note that while the set of states is typically pre-defined and static, the choice of the next conversational move, e.g., whether to ask more questions or show a recommendation, can be determined dynamically, e.g., based on reinforcement learning \citep{christakopoulou2016towards,DBLP:journals/expert/RicciN07}.

In contrast, a typical form of representing dialogue knowledge in a more implicit way is to model the set of supported \emph{user intents}. Such an approach is common in today's chatbot applications and tools like Google's \emph{DialogFlow}. The main task of such dialogue engines is to guess the most probable intent from a user utterance given in natural language. The responses are then commonly constructed by filling templates that are pre-defined for each intent.

End-to-end learning approaches, finally, seek to avoid the knowledge-intensive process of defining explicit dialogue states. These systems aim to learn from data as much as possible how to react appropriately given a current user utterance and the history of the ongoing dialogue. For example, when observing a greeting from the user, the learned model will, after training, respond with a greeting as well.

\paragraph{Discussion}
Our discussions show that the development of CRS comes with many challenges that are not present in ``one-shot'' recommender systems found online or in digital assistants. Determining suitable recommendations given a set of known user preferences, e.g., in the form of previously liked items or explicitly stated preferences regarding item attributes, is probably one of the easier parts. Many existing algorithms for item ranking can in principle be applied. There are, however, many design choices that can be made for the other components. It can, for example, be less than clear which mix of interaction modalities is the best one for a given application. Some users might prefer voice input, others might prefer to type text, and yet other users might have a preference for pre-defined input controls such as radio buttons. Likewise, on the output side, a mix of modalities, e.g., voice and visual outputs, might be desirable by some users, but not all.  Moreover, while providing a human-like avatar or the consideration of additional input signals like facial expressions or gestures might help increase the naturalness of a conversation, such features also add to the complexity of the system.

In the following, we discuss two additional challenges. First, we argue that real-world conversations about recommendations can be rich and multi-faceted, and that a CRS, to be effective, should be able to support such conversations that go beyond simple question-answering (Q\&A) patterns. Second, while the use of learning-based techniques in CRS has the promise to reduce or avoid manual knowledge engineering efforts, the use of machine learning can leave it difficult to make quality guarantees regarding the system's behavior.

\subsection*{Recommendation Dialogues: More than Question-Answering}

Traditionally, the literature differentiates between \emph{system-driven}, \emph{user-driven}, and \emph{mixed-initiative} conversations. In an entirely system-driven approach, the system leads the dialogue, usually by asking questions, e.g., about the user's preferences regarding individual items or item attributes, until it is confident to make a recommendation. Such a guided-dialogue approach is common in knowledge-based and critiquing-based systems. Similar ``system ask---user respond'' dialogue patterns are also found in recent learning-based approaches, e.g., \citep{Zhang:2018:TCS:3269206.3271776}. The other extreme would be an entirely user-driven dialogue, where the CRS only behaves reactively. Voice-controlled smart assistants are an example of such systems that only respond to users, but usually do not take the initiative, e.g., by asking questions back to the users.

Real conversations between humans can however be more varied, e.g., in terms of who takes the initiative and how the dialogue develops. To build a CRS that feels ``natural'', it is therefore important to understand how humans would converse in a given domain. \cite{christakopoulou2016towards} conducted such an analysis in the restaurant recommendation domain as a starting point for their research to understand which questions should be asked to users.

Similar analyses in \citep{CaiChen2020Taxonomy} reveal that, in reality, human recommendation dialogues can be quite rich in terms of what kinds of information are exchanged. Moreover, such information exchanges are often not limited to recommendation-related aspects, such as providing or revising preferences, asking for an explanation, or requesting additional information. In particular, when natural language interactions are supported, we often observe social communication acts such as greetings or phatic expressions. Moreover, besides such human-to-human interactions, there can be specific information exchanges that only happen in human-machine interactions, e.g., when the user wants to restart a broken conversation.

\begin{table*}[t]
        \scriptsize
        \caption{Catalog of user intents}
        \label{tab:user_intent_taxnonmy}
         \setlength{\tabcolsep}{0.6mm}{
        \begin{tabular}{p{3.2cm}p{6.3cm}p{6cm}}
                \hline
                \textbf{Intent} & \textbf{Description} & \textbf{Example}\\
                \hline
                \multicolumn{2}{l}{\textbf{Ask for Recommendation}} \\
                \quad Initial Query & Seeker asks for a recommendation in the
first query. & ``\emph{I like comedy do you know of any good ones?}''\\
                \quad Continue & Seeker asks for more recommendations in the subsequent query. & ``\emph{Do you have any other suggestions?}''\\
            \quad Reformulate & Seeker restates her/his query with or
without clarification/further constraints. & ``\emph{Maybe I am not
being clear. I want something that is in the theater now.'' }\\
                \quad Start Over & Seeker starts a new query to ask for
recommendations. & ``\emph{Anything that I can watch with my kids under
10.}''\\\hline
                \multicolumn{2}{l}{\textbf{Add Details}} \\
                \quad Provide Preference & Seeker provides specific preference
for the item s/he is looking for. & ``\emph{I usually enjoy movies with
Seth Rogen and Jonah Hill.}''\\
                \quad Answer & Seeker answers the question issued by the
recommender. & ``\emph{Maybe something with more action.}'' \newline (Q:
``\emph{What kind of fun movie you look for?}'')\\
                \quad Ask Opinion & Seeker asks the recommender's personal
opinions. & ``\emph{I really like Reese Witherspoon. How about you?}''\\\hline
                \multicolumn{2}{l}{\textbf{Give Feedback}}\\
                \quad Seen & Seeker has seen the recommended item before. &
``\emph{I have seen that one and enjoyed it.}''\\
                \quad Accept & Seeker likes the recommended item. &
``\emph{Awesome, I will check it out.}''\\
                \quad Reject & Seeker dislikes the recommended item. & ``\emph{I
hated that movie. I did not even crack a smile once.}''\\
                \quad Inquire & Seeker wants to know more about the recommended
item. & ``\emph{I haven't seen that one yet. What's it about?}''\\
                \quad Critique-Feature & Seeker makes critiques on specific
features of the current recommendation. & ``\emph{That's a bit too scary
for me.}''\\
                \quad Critique-Add & Seeker adds further constraints on top of
the current recommendation. & ``\emph{I would like something more
recent.}''\\
                \quad Critique-Similar & Seeker requests something similar to the
current recommendation. & ``\emph{Den of Thieves
(2018) sounds amazing. Any others like that?}''\\
\quad Neutral Response & Seeker does not indicate her/his
preference for the current recommendation. & ``\emph{I have actually
never seen that one.}''\\
\hline
            \textbf{Others} & Greetings, gratitude expression, or chit-chat
utterances. & ``\emph{Sorry about the weird typing.}'' \\
                \hline
        \end{tabular}}
        \vspace{-0.1cm}
\end{table*}

\begin{table*}[h]
        \scriptsize
        \caption{Catalog of recommender actions}
        \label{tab:recommendation_action_taxnonmy}
        \setlength{\tabcolsep}{1mm}{
        \begin{tabular}{p{3.2cm}p{6.3cm}p{5.9cm}}
                \hline
                \textbf{Action} & \textbf{Description} & \textbf{Example} \\
                \hline
                \multicolumn{2}{l}{\textbf{Request}}  \\
                \quad Request Information & Recommender requests for the
seeker's preference or feedback. & ``\emph{What kind of movies do you
like?}'' \\
                \quad Clarify Question  & Recommender asks a clarifying question
for more details.& ``\emph{What kind of animated movie are you thinking
of?}'' \\\hline
                \multicolumn{2}{l}{\textbf{Respond}} \\
              \quad Answer & Recommender answers the question asked by the
seeker. & ``\emph{Steve Martin and John Candy.}'' \newline (Q: ``\emph{Who is in
that?}'') \\
  \quad Respond-Feedback & Recommender responds to other
feedback from the seeker.& ``\emph{That's my favourite Christmas movie
too! }''  (U: ``\emph{My absolute favourite!!}'') \\
\hline
                \multicolumn{2}{l}{\textbf{Recommend}} \\
                \quad Recommend-Show & Recommender provides recommendation by
showing it directly. & ``\emph{The Invitation  (2015) is a movie kids
like.}'' \\
                \quad Recommend-Explore & Recommender provides recommendation
by inquiring about the seeker's preference. & ``\emph{Have you seen Cult
of Chucky (2017)  that one as pretty scary.}'' \\\hline
                \multicolumn{2}{l}{\textbf{Explain}}   \\
                \quad Explain-Introduction  & Recommender explains
recommendation with non-personalized introduction. & ``\emph{What about
Sleepless in Seattle (1993)? Hanks and Ryan?}'' \\
                \quad Explain-Preference  & Recommender explains
recommendation based on the seeker's past preference. & ``\emph{Will
Ferrell is also very good in Elf  (2003) if you're in need of another
comedy}'' \\
                \quad Explain-Suggestion  & Recommender explains
recommendation in a suggestive way. & ``\emph{If you like gory then I
would suggest The Last House on the Left  (2009).}''  \\\hline
                \textbf{Others} & Greetings, gratitude expression, or chit-chat
utterances. & ``\emph{Have a good night.}'' \\
                \hline
        \end{tabular}}
        \vspace{-0.1cm}
\end{table*}

\paragraph{A Catalog of User Intents and System Actions}
Several previous studies analyzed the users' intents behind their utterances when they interact with a conversational system in natural language. For instance,
\cite{DBLP:conf/aaai/YanDCZZL17} identified the four most-frequent user intents in a shopping chatbot: \emph{recommendation}, \emph{comparison}, \emph{ask opinion}, and \emph{Q\&A}; and three session-aware intents: \emph{add filter condition}, \emph{see-more}, and \emph{negation}. \cite{kang2017understanding} collected initial and follow-up queries issued by users when they ask for recommendations via speech or text based dialogues. They classified these initial queries into \emph{objective}, \emph{subjective}, and \emph{navigation} goals, and the follow-up queries into the categories \emph{refine}, \emph{reformulate}, and \emph{start over}.

Most recently, \cite{CaiChen2020Taxonomy} established a catalog of user intents based on an analysis of human-human dialogues centered around movie recommendations. Specifically, they annotated a subset of conversations from the ReDial dataset after performing a cleaning process. The final catalog of user intents was then created using an iterative \emph{grounded theory} procedure. In the catalog, the intents are classified into three top-level intents, i.e., \emph{Ask for Recommendation}, \emph{Add Details}, and \emph{Give Feedback}, and 15 sub-intents (see Table \ref{tab:user_intent_taxnonmy}). For instance, there are four sub-intents under \emph{Ask for Recommendation}, including \emph{``Initial Query''}, \emph{``Continue''} (continuing to seek for more suggestions), \emph{``Start Over''} (when the recommendation seeker starts a new query), and \emph{``Reformulate''} (when the seeker wants to revise a previous query).

Moreover, \cite{CaiChen2020Taxonomy} classified the actions of the human recommenders into four top-level categories, i.e., \emph{Request}, \emph{Respond}, \emph{Recommend}, and \emph{Explain}, and nine sub-actions (see Table \ref{tab:recommendation_action_taxnonmy}).
An analysis regarding the action distribution shows that the sub-actions under \emph{Explain} more frequently occur in satisfactory dialogues, i.e., in dialogues where at least one recommended item was liked by the seeker. This may imply that providing explanations for the recommendation is likely to increase user acceptance.

\paragraph{Discussion}
While logged dialogues between humans represent a valuable resource for understanding how humans interact, crowdsourced datasets like ReDial have their limitations. When the ReDial data was collected, the crowdworkers were required to mention at least four movies during their conversation. This might have led to dialogues that mostly focus on the instance level. As a result, preferences are often obtained by asking for the seeker's general tastes or opinions on specific movies. Possible other types of preference elicitation---such as asking for pairwise preference feedback on two or more items \citep{RanaBridge2020}---rarely occur.

Regarding the developed catalog of user intents and system actions, note that the analysis was done for only one domain (i.e., movies), and that additional user intents might be relevant in other domains. Also, investigating chit-chat intents in more depth seems advisable, e.g., in terms of \emph{in which condition} users might have such intent, and \emph{what kind of chit-chat} they may expect to have. In a recent work, \cite{liu-etal-2020-towards} developed a CRS to cover multiple dialogue types such as \emph{Q\&A}, \emph{chit-chat}, and \emph{recommendation} in a primarily system-driven approach, where the system proactively leads the conversation by following a planned goal sequence. However, it seems that their collected dialogue dataset (called DuRecDial) was limited to pre-defined user profiles and task templates, which might not fully reflect the characteristics of natural conversations.

Generally, traditional CRS such as critiquing approaches are mostly limited to form-based preference elicitation and in many cases based on a predefined set of attributes. Today's CRS often only support a smaller subset of user intents. 
Some of end-to-end learning-based systems, on the other hand, sometimes only support a limited set of actions like asking for a preference and recommending an item, and they are unable to provide explanations or consider user preferences about item attributes.  Overall, more research is therefore required to better understand how humans interact in recommendation dialogues, in particular in terms of the social communication acts that may help to increase the acceptance of recommendations \citep{hayati2020inspired}.

\subsection*{The Need for Predictability \& Quality Guarantees}

According to our discussions, there are two extreme approaches of how to build the core of a CRS. One extreme is to explicitly encode the different types of knowledge needed for dialogue management and item recommendation, leading to an entirely deterministic system. Such an approach is desirable or even required when the quality of the responses must be guaranteed, e.g., when the system makes recommendation in the health or finance domains. The other extreme is to try to learn everything from observed dialogues between humans, i.e., from mostly unstructured data. Various existing approaches are placed somewhere between these two extremes \citep{DBLP:journals/expert/RicciN07,christakopoulou2016towards}.

Explicitly modeling the dialogue and recommendation knowledge, e.g., in the form of state automata and recommendation rules, can require substantial knowledge engineering. Special-purpose modeling environments may  help to reduce these efforts \citep{Jannach:2004:ASK:3000001.3000153}. However, the problem remains that the knowledge bases have to be manually updated, e.g., when new items become available that require the change of the recommendation logic or the revision of the dialogue flow.
Also, these systems are by design restricted in terms of which interactions they support. For example, free-text inputs in natural language are usually not supported.

One ultimate vision of a pure end-to-end learning system, as the other extreme, is that the behavior of the CRS (e.g., the choice of dialogue moves and the recommendations) is entirely learnt from data. In such an approach, no knowledge engineering is required and the system automatically updates its models whenever new data arrive. At the same time, such learning-based systems are not limited to a predefined set of dialogue situations and may be able to react to a rich set of user intents that were observed previously in the data.

One main challenge of learning-based systems, however, is that there are usually no quality guarantees regarding the system responses, as  they are not entirely predictable. One main problem is that there are several places in a typical CRS architecture where a learning-based system can fail. During natural language input processing, for example, such systems may encounter difficulties in speech-to-text conversion, named entity recognition, or intent detection. Given these uncertainties, it might therefore be desirable to constrain the possible outputs of a learning-based CRS to assure a certain level of predictability and guaranteed output quality. This may start with constraints at the grammatical level when generating sentences with the aim to avoid ungrammatical system responses. At the dialogue level, one may furthermore constrain the system to a predefined set of user intents and system actions. Finally, one could also restrict the final outputs to be based on predefined and quality-assured response templates, which turns response generation into a retrieval task. All these measures may ultimately help to increase the predictability (and quality) of the system output. However, these measures also add several knowledge-based components to the overall architecture, leading to challenges regarding knowledge modeling and maintenance.

To exemplify the challenges of today's learning-based systems, consider the case of the approaches by \cite{li2018towards} (termed DeepCRS) and \cite{chen-etal-2019-towards} (termed KBRD), which were presented in recent years at leading conferences on natural language processing and neural networks. Both approaches are based on the ReDial dataset, and KBRD also relies on structured external knowledge. An evaluation of these systems by \cite{JannachManzoor2020} indicated that both systems seem to make too many mistakes when generating responses. For both systems, at least 30\% of the system responses were considered to be not meaningful dialogue continuations according to the authors. Typical problems include that the system did not properly react to a user question, repeatedly made the same recommendation, or abruptly ended the dialogue.
These findings raise questions regarding the usefulness of such unconstrained academic approaches in practice. An interesting side observation in the mentioned evaluation was that both  end-to-end systems actually did not generate new sentences and almost all of the returned responses appeared in similar or identical form in the training data.

Clearly, our general goal when building CRS of the future is not to rely on pre-implemented dialogue flows and knowledge-based recommendation systems, but to design ``intelligent'', learning-based solutions. More work is thus needed to understand how useful our solutions are in practice and how we can build systems that guarantee a certain quality level . Moreover, better methodological approaches seem to be needed to assess the quality of a CRS in the lab and how to compare different conversational systems. We discuss such methodological questions in the next section.

\section*{Why Evaluating a CRS is Challenging}

\paragraph{CRS are Multi-Component Adaptive Interactive Systems}
Modern CRS are in general complex, multi-component software systems, which have to support non-trivial interactions with their users. Consequently, the quality perception and the adoption of such systems by users can depend on a variety of factors, and in case the system is not as successful as expected, many components may be the culprits. The system might, for example, have failed to understand the user utterance or to extract the correct intent. Or, it might have not been able to respond appropriately to question, or it might just have made poor recommendations. Finally, there might have been issues with the user interface or the supported interaction modality.
Overall, any evaluation of a CRS therefore must aim to isolate these potential factors to see what works well and what does not.

In that context, note that the evaluation of the quality of the underlying \emph{recommendation algorithms} alone can be challenging in its own. Today, the research community largely relies on offline evaluation procedures using historical datasets to assess how good an algorithm works. Such an approach can be suited to assess how good an algorithm is at predicting which item a user will rate highly or click on. However, these evaluation procedures cannot inform us about the quality perceptions by users, for instance, if the recommendations will help users discover new items, or if the recommendations will lead to higher business value. Such aspects can only be assessed through user studies or field tests. In fact, in CRS, the majority of the aspects that may contribute to the success or failure of the system cannot be reasonably assessed without involving humans users. In contrast, both in earlier critiquing-based CRS and in recent chatbot applications a very common approach is to assess the number of required interaction turns until the user finds a suitable recommendation through simulations. Such simulations however rely on a number of assumptions like rational user behavior or that the user preferences are stable and pre-existing. Moreover, in practice, longer interaction sessions with a CRS not necessarily mean that the system does not work well. Instead, longer interactions might in contrast indicate higher engagement and more exploration by users \citep{Cai:iui2021}.

\paragraph{The Need for Multi-Faceted Evaluations}
Generally, we can identify three main dimensions in which CRS are evaluated:
\begin{enumerate*}[label=\emph{(\roman*)}]
\item \emph{Effectiveness of Task Support},
  \item \emph{Efficiency of Task Support}, and
  \item \emph{Quality of the Conversation and Usability}.
\end{enumerate*}
Ultimately, all of these aspects can contribute to the success of a CRS in practice.

\emph{Effectiveness of Task Support}: These measurements relate to the ability of the CRS to support a recommendation-related task, e.g., helping users to make a decision or to find an item of interest. In the literature, researchers assess such aspects with the help of evaluations with users and /or through offline experiments. In user studies, often both \emph{objective} and \emph{subjective} measures are applied. Objectively, one can for example determine how often users actually found an interesting item during the study. This can, for example, be measured by counting add-to-cart actions or \emph{task completion rates}. Common subjective (self-reported) measures in user studies include \emph{decision confidence}, \emph{perceived recommendation quality}, and \emph{purchase/return intentions}. A more general level, \emph{user satisfaction}---usually with the system as a whole---is frequently used in the literature as well.

In offline studies on effectiveness, common proxies from non-conversational systems are typically applied, including all types of accuracy measures like Precision, Recall and RMSE. Often, such measures are taken in a simulation based approach, where artificial users with pre-defined preferences interact with the CRS. Besides accuracy, some studies also measure the \emph{success rate} and \emph{rejection rate} in simulations. Since any CRS is a multi-component system, researchers sometimes focus on individual components in their evaluation. A typical problem, for example, is to assess the performance of the entity and intent recognition modules as in \citep{liao2019deep} or \citep{Narducci2018aiai}.

\emph{Efficiency of Task Support}: Measurements in this category assess how quickly users make a decision or find something suitable. As discussed above, one underlying assumption is that shorter dialogues are preferable. Traditionally, the number of \emph{required interaction cycles} in simulated user-machine conversations is measured. An alternative objective measure is \emph{task completion time}.
\cite{iovine2020conversational}, for example, compared the use of different interaction modalities of a CRS---language-based, button-based, mixed---in a user study. They found that natural language interfaces can lead to a less efficient recommendation process, which is partially due to the challenge of understanding the natural language input.  The use of voice-based interactions was studied in \citep{Yang:2018:UUI:3240323.3240389}. Here, the authors found that users of a podcast recommender were slower and explored fewer options when interacting through voice. Combined, these findings indicate that natural language based interaction modalities may make the interaction process less efficient than when buttons and forms are used.

\emph{Quality of the Conversation and Usability}: Commonly, these aspects are evaluated through subjective measures, where users are asked about their quality perceptions. In terms of general usability, the \emph{ease-of-use} of the system or the \emph{task ease} are often in the focus. Looking at the quality of the recommendation process itself, researchers furthermore addressed questions of \emph{transparency} or the perceived level of \emph{user control}. At the dialogue level, researchers furthermore applied ideas that were previously applied for general spoken dialogue systems, and measured how quickly a system adapts to the user's preferences, how intuitive and natural the dialogue feels, and if the dialogue is entertaining. Moreover, \cite{Pecune:HAI2019} considered \emph{coordination}, \emph{mutual attentiveness}, \emph{positivity}, and \emph{rapport} as factors that might affect the perceived dialogue quality. In other approaches, a variety of additional subjective measures were used, including \emph{consistency}, \emph{engagingness}, \emph{informativeness}, or \emph{relevance}.

Recently, various attempts were also made to \emph{objectively} measure linguistic aspects of the system's generated utterances. One idea is to compare these utterances with ground-truth utterances by humans in a given dialogue situation, using, e.g., the BLEU or NIST scores, which are commonly used in machine-translation tasks. Alternative linguistic measures include the lexical diversity or perplexity (fluency), e.g., in \citep{chen-etal-2019-towards}.

\paragraph{Discussion}
The recommender systems research community has developed broadly accepted standards regarding the evaluation of such systems. In the predominant area of algorithms research, offline experimentation and the use of metrics from information retrieval and machine learning are common. On the other hand, for user-centric research, different general evaluation frameworks were proposed, e.g., by \cite{Pu:2011:UEF:2043932.2043962} or \cite{umuai2012knijnenburg}. Both types of evaluation approaches can be applied for CRS as well. The general limitations of offline evaluations however remain, i.e., that it is not always clear if offline results are representative of the user-perceived qualities of the recommendations. The user-centric frameworks, on the other hand, are by design focusing on general aspects of recommender system acceptance such as the perceived recommendation quality, usability aspects, or the intention to use the system in the future.

In the context of CRS, given their interactive nature, offline evaluations might only be informative for very specific subtasks, such as the recognition rate of entities in user utterances, which are often not very specific to CRS. As a result, almost all research efforts in some way might require the involvement of humans in the evaluation process. Recent works on end-to-end learning approaches, as those mentioned above by \cite{li2018towards} (DeepCRS) and \cite{chen-etal-2019-towards} (KBRD), therefore apply a combined approach in their evaluations, which is based both on objective measures in offline evaluations and on human assessments of the dialogue quality.

There are, however, a number of potential pitfalls to be considered. For example, using metrics like the BLEU score in offline experiments to compare a system-generated response with a ground-truth response from a recorded dialogue may be problematic in different ways. First, the commonly-used BLEU score, according to \cite{liu-etal-2016-evaluate}, not in all cases seems to correlate with human perceptions. Second, an evaluation procedure where the system response is compared to a ground-truth statement might be too limited, because many alternative generated statements might be appropriate as well in a given dialogue situation. Moreover, remember that in some works, metrics are used to measure the fluency of the system responses. However, in cases where the end-to-end learning system is actually not generating new responses, but only returns utterances that appear in the training data, the application of fluency metrics might not be too meaningful.

User-centric research can also have pitfalls. When evaluating the DeepCRS and KBRD system, the authors for example rely on human annotators to rank the responses by different systems in a given dialogue situation, e.g., in terms of the consistency with what was said before in the dialogue. However, in some evaluations, the task of the annotators was to assess which system response is better. Given only such \emph{relative} judgments, it unfortunately remains unclear if the systems generate high-quality responses on average on an \emph{absolute} scale. Moreover, for the evaluation of some specific aspects of CRS, e.g., regarding the naturalness or engagingness of the conversation, very specific experimental setups are required. Unfortunately, no commonly accepted evaluation standards for such questions exist so far.

\section*{Moving Forward} 
\paragraph{Improving Our Research Methodology}
A direct consequence of our discussions is that we should aim at further improving our evaluation methodology. Such improvements could for example be achieved through the development of a unified user-centric evaluation framework for CRS, which
\begin{enumerate*}[label=\emph{(\roman*)}]
\item considers and relates the various factors that may influence the quality perceptions of users, and
\item provides standardized questionnaire for the different model constructs; see also \citep{Pu:2011:UEF:2043932.2043962}.
\end{enumerate*}
Moreover, in case of offline evaluations, the current metrics that we use to automatically assess the linguistic quality of system-generated responses have to be re-assessed and validated.

Generally, it seems advisable that we rely on a richer methodological repertoire, which allows us to conduct multi-modal and multi-metric evaluations to obtain a more comprehensive picture of the quality of the different aspects of a CRS. In that context, also more \emph{exploratory} research is needed, for example, to understand how humans interact in conversations, what users expect from a computerized advisor, or how tolerant they are with respect to problems in such a conversation.

\paragraph{Combining Learning and Knowledge Based Systems}
Our analysis of recent end-to-end learning system indicates that building a CRS solely based on learning from logged interactions still has a number of limitations. One reason for this could lie in the current lack of datasets that contain richer interactions between humans. The ReDial dataset, as mentioned above, in some ways may appear unnatural. This seems to have led to the effect that mentions of desired movie attributes like genres or actors are under-represented, which makes learning difficult.

Therefore, for the time being, hybrid AI-based solutions seem to be the method of choice, where certain parts of the needed knowledge are learned from past data, certain parts are represented in structured form and taken from external sources such as DBPedia, and certain parts are manually engineered based on domain expertise. How to combine learned and explicit knowledge in the best way is a very active research area in AI. Moreover, as discussed above, more research is required to understand how to provide certain guiding rails for learning based systems in order to ensure predictable, high-quality
responses.

\paragraph{Supporting Novel Interaction Forms and Application Domains}
In their traditional form, CRS were mostly web-based applications with a pre-defined form for eliciting user feedback on a set of fixed attributes. Due to recent technological advances, natural language input, either in written or spoken form, has become mainstream. In the future, a more frequent use of additional or alternative forms of interactions can be envisioned. In the literature, we, for example, find approaches that consider \emph{non-verbal} communication acts like body postures, gestures, and facial expressions as inputs. But also new forms of output and feedback are possible, in particular by using novel forms of \emph{Embodied Conversational Agents} or application-specific 3D visualizations.

Future CRS might also not be limited in terms of the application environment. Today, CRS are mostly implemented as desktop or mobile applications. Increasingly, we also see CRS functionality implemented on smart speakers like Amazon Echo. Very differently from that, future CRS might also be part of other physical environments, e.g., in the form of an interactive wall installed in a real shop, a service robot in a restaurant, or an in-car system.

\paragraph{Towards Social CRS}
An analysis of the recorded human-human conversations in the ReDial dataset shows that a substantial fraction of the dialogue represents \emph{phatic} utterances like chit-chat.  To further increase the adoption of chatbot-like CRS, it might therefore be highly important to further enhance such a functionality so that the system is able to establish a social connection to the users. In \citep{FromEliza2018}, for example, the authors report that users of a social chatbot by Microsoft, which is able to detect the users' emotional needs and personalities, even possess a certain feeling of ``social belonging''.

While today's end-to-end learning systems seem able to engage in chit-chat conversations, limited work exists so far on understanding the users' current emotions or to react on them. One reason for the lack of such capabilities may lie in the characteristics of many of today's datasets, which do not reflect the full spectrum of how humans would engage in real-world conversations. Recently, a new dataset called INSPIRED was released by \cite{hayati2020inspired}, and the authors show that an end-to-end learning model trained on the data annotated with social strategies can be beneficial. Future work might build on these results and the released datasets to build next-generation social CRS. For instance, as summarized by \cite{thomas2020theories}, a number of human-human conversation strategies (such as social norms, structures, affect, prosody, and style) might be leveraged to enable the system to be more engaging, persuasive, and trustworthy.

\section*{Summary}
Being able to conduct natural conversations with a human-like computerized system is a long-standing vision of AI. We have reviewed existing approaches to building interactive advice-giving systems in the form of conversational recommender systems. Our discussions indicate that today often a trade-off exists between mostly engineered and entirely learning-based solutions. While engineered solutions may excel in terms of predictability and quality guarantees, learning-based systems have the promise to support much more natural and flexible conversations.

\bibliography{acmart}

\end{document}